\documentclass{article}

\usepackage[preprint]{neurips}

\usepackage[utf8]{inputenc} 
\usepackage[T1]{fontenc}    
\usepackage{url}            
\usepackage{booktabs}       
\usepackage{amsfonts}       
\usepackage{nicefrac}       
\usepackage{microtype}      
\usepackage{xcolor}         
\usepackage{listings}
\usepackage{ulem}
\usepackage{enumitem}
\usepackage{tabularx}
\usepackage[most]{tcolorbox}
\usepackage{adjustbox}
\usepackage[hidelinks]{hyperref}

\hypersetup{colorlinks=true, allcolors=blue}
\definecolor{panelbg}{HTML}{E9E9E9}
\definecolor{headerbg}{HTML}{4E4E4E}

\lstdefinestyle{gsmcode}{
    backgroundcolor=\color{panelbg},
    basicstyle=\ttfamily\small,
    columns=fullflexible,
    keepspaces=true,
    showstringspaces=false,
    breaklines=true,
    frame=none,
    xleftmargin=0pt,
    xrightmargin=0pt
}

\usepackage{xcolor}

\title{AIMO Interpretability Challenge}

\author{
    \!\!\!\!\!\!\!\!\!\!\!\!Michal Štefánik$^{1*}$\ \ \  Philipp Mondorf$^{2*}$\ \ \ Andreas Waldis$^{3*}$\ \ \ Michal Spiegel$^{5*}$\ \ \ Robert Belanec$^{10,11*}$\\\vspace{3pt}$^{*}$core organisers\\ \textbf{Chuan Yang}$^{4\psi}$\ \ \ \ \textbf{Maria Matveev}$^{12\psi}$\ \ \ \ \textbf{Jan Mazak}$^{13\psi}$\ \ \ \ \textbf{Qianying Liu}$^{1\psi}$\ \ \ \ \textbf{Markus Schepers}$^{14\psi}$\\$^{\psi}$expert mathematicians\vspace{3pt}\\ \textbf{Josef Kuchař}$^{5\omega}$\ \ \ \ \textbf{Marek Kadlčík}$^{5\omega}$\ \ \ \ \textbf{Adam Vawda-Oomerjee}$^{6,1 \omega}$\ \ \ \ \textbf{Chaoran Liu}$^{1\omega}$\\$^{\omega}$infrastructure team\vspace{3pt}\\\textbf{Simon Frieder}$^{7,8  \phi}$\ \ \ \textbf{Barbara Plank}$^{2 \phi}$\ \ \ \textbf{Fazl Barez}$^{8,9 \phi}$\ \ \ \textbf{Pontus Stenetorp}$^{6,1 \phi}$\\$^{\phi}$advisory board\vspace{10pt}\\
    \url{https://aimo-interp.github.io}\vspace{-10pt}
}
\begin{document}
\footnotetext[1]{National Institute of Informatics, Japan\ \ \ 
$^2$Munich Center for Machine Learning / MaiNLP LMU \\
$^3$University of Tübingen\ \ \ 
$^4$Fuzhou University\ \ \ 
$^5$Masaryk University\ \ \ 
$^6$University College London \\
$^7$AIMO Organization Team\ \ \ 
$^8$University of Oxford\ \ \ 
$^9$Martian
$^{10}$Faculty of Information Technology, Brno University of Technology\ \ \ 
$^{11}$Kempelen Institute of Intelligent Technologies\ \ \ 
$^{12}$LMU Munich\ \ \ 
$^{13}$Comenius University\ \ \ 
$^{14}$Johannes Gutenberg University Mainz
}

\setcounter{footnote}{0}

\maketitle

\begin{abstract}
We propose the AIMO Interpretability Challenge, a competition on distinguishing \emph{robust} from \emph{spurious} reasoning in frontier mathematical language models based on the models' internal mechanisms. The challenge is motivated by a central limitation of standard reasoning benchmarks: strong final-answer accuracy does not reveal whether a model relies on stable reasoning mechanisms or exploits brittle reasoning shortcuts. Building on AI Mathematical Olympiad (AIMO) problems and submissions, together with resources from the Fields Model Initiative, the competition will provide (1)~newly-published olympiad-level math reasoning problems and their symbolic representations, allowing generation of novel functional variants, (2)~access to frontier reasoning models, and (3)~assessments of models' adversarial robustness on these problems. Participants will use these resources, along with our computing infrastructure support, to develop methods for identifying which models solve problems robustly.
Our competition will also create a new, open robustness benchmark and baseline systems, aiming to provide a lasting foundation for standard benchmarking in mathematical reasoning and interpretability. 
Scientifically, the competition connects interpretability and generalization research around a central question in AI research: can we determine if, and to what extent, the decision-making of frontier AI models is generalizable and thus, reliable?
\end{abstract}


%
%
\section{Competition description}

\subsection{Background and impact}
Recent advances of large language models have sharpened a foundational disagreement in AI: whether frontier systems are beginning to exhibit generalized, frontier reasoning abilities, or whether they remain highly capable pattern matchers whose success often depends on spurious cues \citep{feng2024how,mondorf2024beyond,mirzadeh2025gsmsymbolic}. Standard benchmark scores alone can not resolve this disagreement, because they can not evidence \emph{how} a system arrives at a correct answer. Interpretability offers a promising route forward: previous work uncovers robust mechanisms of internal representations \citep{stefanik2025unravellingmechanismsmanipulatingnumbers} or identifies circuits claimed to implement meaningful functions decomposing the high-level problem into subproblems \citep{wang2023interpretability,brinkmann-etal-2024-mechanistic} -- implying that
models' underlying mechanisms are indeed vastly robust. However, much of current work is still dominated by compelling case studies and localized analyses that are difficult to compare and generalize \citep{bereska2024mechanistic,templeton2024scaling,lindsey2025biology}.

The proposed competition targets this gap. Its scientific goal is to measure whether interpretability methods can identify mechanisms that generalize across problem instances and meaningfully distinguish robust models from brittle ones. This problem spans multiple NeurIPS-relevant areas, including interpretability, evaluation, generalization, reasoning, mathematical AI, and AI safety.

The submissions of this competition will be applicable in auditing high-stakes reasoning systems. In realistic deployment, a research lab, education provider, or scientific-assistance platform may choose among several models that achieve similar benchmark accuracy, yet differ substantially in reliability. The competition operationalizes this scenario by asking participants to infer robustness from behavior and internal model representations.
Beyond the competition itself, the resulting symbolic robustness benchmark will also support future work focused on developing more robust frontier models.

We expect interest from both a general ML audience interested in models' frontier capabilities, and the interpretability community interested in understanding the models' internals. In 2026, AIMO 3\footnote{\url{https://aimoprize.com}} that was hosted  Kaggle\footnote{\url{https://www.kaggle.com/competitions/ai-mathematical-olympiad-progress-prize-3}} attracted more than 4{,}000 teams, indicating substantial upstream interest of the AI community in olympiad-level reasoning systems. In our challenge, we expect roughly 20--40 well-performing submissions from teams of researchers focused on interpretability and generalization.

\subsection{Novelty}


With regard to prior competitions at ML conferences, the AIMO Interpretability Challenge is an entirely new competition. Its novelty lies in connecting interpretability with an objective that is increasingly relevant for the broader AI community, but standard benchmarks do not directly target: assessing whether frontier models \emph{that succeed} on most complex reasoning tasks do so \emph{robustly}. 

While this competition will not build the first interpretability benchmark available, the benchmark developed in this challenge will also make a valuable and lasting contribution to the field of interpretability.
Existing interpretability benchmarks focus on evaluating interpretability methods in rudimentary and/or synthetic settings; Most aligned with us, the RAVEL benchmark tests whether methods can identify and disentangle attribute-value information \citep{huang-etal-2024-ravel}, InterpBench tests whether methods recover planted internal algorithms from semi-synthetic transformers with known circuits \citep{gupta2024interpbench}, and MIB advances standardized evaluation of circuit and causal-variable localization on synthetic, controlled tasks such as arithmetics \citep{pmlr-v267-mueller25a}.

These benchmarks present valuable contributions for developing better interpretability methods, but their highly controlled settings limits their contribution in our understanding of the behaviour of frontier models. Our goal is not to validate recovery of known subcomponents, but to test whether interpretability can determine whether strong performance on hard reasoning problems reflects stable, generalizable reasoning mechanisms or brittle strategies that fail under structured variation.
This goal is aligned with some recent work striving to identify generalization from internal circuitry \citep{huang2025internal} or attention patterns \citep{li2025interpretationpredictbehaviorunseen,spiegel2025attendperishbenchmarkingattention}.
To this line of work, the proposed competition brings both a standard benchmark and a reusable toolset for understanding the nature and limits in frontier reasoning capabilities and with the most advanced models.

%
%
%
\subsection{Data}

\begin{figure}[t]
\centering
\vspace{-0.5em}
\begin{tcolorbox}[
    colback=panelbg,
    colframe=headerbg,
    boxrule=0.6pt,
    arc=3pt,
    outer arc=3pt,
    width=\linewidth,
    title=\textbf{AIMO 2 Reference problem 480182},
    coltitle=white,
    colbacktitle=headerbg,
    enhanced,
    left=4pt,
    right=4pt,
    top=3pt,
    bottom=3pt
]
\begin{lstlisting}[style=gsmcode,basicstyle=\ttfamily\scriptsize,columns=fullflexible,keepspaces=true,aboveskip=1pt,belowskip=0pt]
PROBLEM: Let ABC be a triangle with BC=[input1], CA=[input2], and AB=[input3]. Point X lies on segment 
         AC such that BX bisects angle CBA. Let omega be the circumcircle of triangle ABX. Let Y be 
         a point on omega different from X such that CX=CY. Line XY meets BC at E. The length of the 
         segment BE can be written as m/n, where m and n are coprime positive integers. Find m+n.
INPUT CONSTRAINTS: CA > BC >> AB (BC > AB * 2)

def solution(bc, ca, ab):
    cx = (bc * ca) / (bc + ab)          # Step 1: Find CX by applying the Angle Bisector Theorem.
    power_c = cx * ca                   # Step 2: Compute the power of C wrt the circumcircle of ABX.
    cb_prime = power_c / bc             # Step 3: Find the circle's second intersection with line BC.
    x = (bc * cb_prime + cx * cx) / (bc + cb_prime)  # Step 4: Locate E on BC by equating powers.
    be = abs(bc - x)                    # Step 5: Convert E's coordinate into segment length BE.
    best_num, best_den = min(           # Step 6: Approximate BE as a fraction m/n.
        ((round(be * den), den) for den in range(1, 1000)),
        key=lambda t: abs(be - t[0] / t[1]))
    a, b = best_num, best_den           # Step 7: Reduce m/n using the Euclidean algorithm.
    while b:
        a, b = b, a % b
    return best_num // a + best_den // a
\end{lstlisting}
\end{tcolorbox}
\vspace{-0.8em}
\caption{Newly annotated symbolic template for AIMO 2 reference problem 480182.}
\vspace{-0.5em}
\label{fig:symbolic_chain}
\end{figure}




The competition uses three main data components:
\begin{enumerate}[leftmargin=1.5em]
    \item \textbf{Original olympiad-level mathematical problems} The core benchmark will be drawn from frontier mathematical problems included in AIMO 2, AIMO 3, JMO (Japanese Mathematical Olympiad), and two last years of AIME (American Invitational Mathematics Examination), totaling 180 problems. Thanks to the close involvement of mathematicians in our organizing team, we can guarantee that 50\% of problems in both the train and test sets are \emph{new}, i.e. not previously published on the Internet. We have confirmed that the licenses for these collections permit their use in the competition (public AIMO and AIME problems are both Apache 2.0-licensed, and we have received confirmation from the owners of the private AIMO and JMO problems). 
    \item \textbf{Symbolic reasoning annotations} For the original problems, we create new annotations of symbolic reasoning chains similar to the functional variation introduced by GSM-Symbolic~\citep{mirzadeh2025gsmsymbolic} but centered on the frontier reasoning level: these chains capture the logical inference underlying valid solutions, allowing us to generate unconstrained volumes of \emph{adversarial problem variants} following controlled distribution shifts (Fig.~\ref{fig:symbolic_chain}). We will use these to identify and annotate the interpreted models for their robustness.
    In this collection, we build on deep expertise in mathematics brought by a direct involvement of olympiad-level mathematicians in our organizational board, including mathematicians from Fuzhou university in the PRC, National Institute of Informatics in Japan and their network of frontier mathematicians at Peking university. 
    By the submission date, we are close to completing the annotation process and will have the full collection ready within one month from the submission deadline.\footnote{A sample of public problems with our newly annotated, verified symbolic chains allowing perturbations is available on: \url{https://huggingface.co/datasets/aimo-interp/problems-public}}
    The full train split of problems with associated symbolic reasoning chains will be publicly released for any use under permissible Apache 2.0 license before the competition starts.
    
    In Appendix~\ref{appx:adversarial_evaluations}, we provide results of robustness to permutations with already-collected symbolic chains, evidencing that they indeed allow us to generate problem variants that can systematically mislead frontier AI models within and beyond AIMO, including Qwen, GPT 5.2, and Gemini-Pro.

    \item \textbf{Interpreted model set} As the analysed models, the challenge will pick among the top-performing AIMO submissions and derive robust/spurious labels from their evaluation on the counterfactual evaluation set.
    Participation of AIMO organisers in this challenge provides competition participants and us with easy access to these models.
\end{enumerate}

\textbf{Data splits\ \ \ } The collected problems will be split into the train and test collections:
\begin{itemize}
    \item \textbf{Train split} will include 150 problems from both published and not-yet-published sources~(AIMO, AIME, JMO).
    \item \textbf{Test split} will contain 30 problems provided by the AIMO 3 organizers at varying levels of difficulty. These problems are and will not be made public and will only be accessible by the participants' submissions through automated evaluation on our servers.
\end{itemize}

The data volume is sufficient for training and a meaningful evaluation because the effective size of robustness labels (Sec~\ref{sec:tasks}) comes from the cross-product of original problems and models. We aim to include at least 8 top-performing models in our evaluations, resulting in roughly 1200 training and 240 test instances.
Nevertheless, to support more data-demanding methods, we will additionally release an LLM-based codebase to generate adversarial training data from generic perturbations not requiring symbolic chains.
Our new benchmarks will not involve personal or sensitive user data. New symbolic chains are created directly by the organizers rather than collected from human subjects.


%
%
\subsection{Tasks and application scenarios}
\label{sec:tasks}

The central task of the AIMO Interpretability Challenge is to determine which model provides a correct answer to a given problem robustly. Formally, given a triple of 
$$(\text{Problem } X, \text{Model } M, \text{Distribution shift } \Delta)$$
where $M(P)$ is correct (i.e. $M(X) = y_\text{true}$), the task is to determine whether $M(\Delta X)$ will also be correct for all perturbations under $\Delta$.
The participants will be asked to create a submission system $S$ that assigns one of two categories to each triple $(X, M, \Delta)$:
\begin{equation*}
S(X, M, \Delta) =
\begin{cases}
0 & \text{if for a majority of } \Delta X \in \Delta: M(X) = y_\text{true} \text{ and } M(\Delta X) \neq y_\text{true} \\
1 & \text{otherwise}
\end{cases}
\end{equation*}

\begin{table}[t]
\centering
\vspace{-0.8em}
\begin{tcolorbox}[
    top=2pt,
    bottom=2pt,
    colback=yellow!20!white,
    colframe=yellow!70!black,
    title=Challenge summary,
    width=\linewidth
]
Given a mathematical problem and a model providing a correct solution to this problem, the task is to decide if the model responds to the given problem \textit{robustly} w.r.t. to a pre-defined set of distribution shifts.
\end{tcolorbox}
\vspace{-2.0em}
\end{table}

While determining a general robustness of $M$ across all possible distribution shifts is intractable, annotations of symbolic chains (Fig.~\ref{fig:symbolic_chain}) offering an unconstrained number of perturbed problems $\Delta X$ will allow us to generate a sufficiently large and broad set to uncover a systematic robustness or fragility of $M$ under $\Delta$.
This abundance will also allow us to leave out from evaluation sets the samples that are not clear-cut; i.e. where either the original model is not robustly correct (e.g. in a repeated sampled generation) or the evaluated perturbations $\Delta X$ do not cause a consistent deterioration to the $M$'s prediction.

The distribution shifts $\Delta$ will be instantiated by applying one or more perturbation types on the original problem using its symbolic representation. A non-exhaustive list of perturbations that can be instantiated using our symbolic chains and that we already tested can be found in Appendix~\ref{appx:adversarial_evaluations} together with examples of their corresponding performance drops.


The challenge is offered in two tracks differing in the scale of analyzed models:
\begin{itemize}[leftmargin=1.5em]
    \item \textbf{Main track} including a full scale of top-performing models from AIMO 3;
    \item \textbf{Small models track} subsetting the evaluation to best-performing models below the 10-billion-parameter scale, providing a comparable setup for methods that are more compute-heavy to train, such as Sparse Autoencoders or Transcoders. 
\end{itemize}

\paragraph{Application}
The competition task corresponds to a concrete real-world need: selecting and auditing the robustness of strong reasoning models before deployment. In industry, this matters for systems that support scientific work, quantitative analysis, or educational tutoring, where a model that appears strong on a benchmark but fails under innocuous reformulations may have costly or harmful consequences. 
Nevertheless, our competition is primarily aimed at contributing to support a central, open research goal: understanding whether, and to what extent, models' successes in tasks of frontier complexity stand upon robust reasoning mechanisms or brittle shortcut following.

\paragraph{Feasibility}
The presented challenge is scientifically challenging but tractable; 
It is challenging because models' black-box behaviour on the original problem will in most cases likely not be sufficient for dissecting robust and non-robust inference mechanisms,\footnote{While this is just an assumption, we note that showing the contrary would have significant implications for a reliability of deployments, itself being a valuable contribution.} reducing the usefulness of standard evaluation and making the distinction to benefit from deeper behavioural or mechanistic evidence. At the same time, it is tractable because the competition provides a clear extrinsic signal of success -- providing labelled datasets that support experimentation with a wide range of possible methods and strategies.

\paragraph{Code of Ethics}
The competition is aligned with the NeurIPS Code of Ethics; its primary aim is to improve reliability assessment of AI systems, it does not target or profile people and it supports safer deployment by helping practitioners identify brittle systems before they are used in consequential settings. Potential risks include overclaiming the strength of interpretability methods based on the competition results. We will mitigate these risks through shaping the submission rules against overfitting, maintaining a private test set, and open-sourcing the full scope of our evaluations.

%
\subsection{Metrics}
The primary metric will be \textbf{accuracy} on the held-out test set, i.e., the fraction of examples for which a submission correctly classifies the model as robust or non-robust.
This metric directly reflects upon our main research question of whether we can dissect  the robustness of LLMs decision-making mechanisms by their internal behaviour.

Together with the global accuracy, in our final report, we will also provide evaluations dissecting performance: (1) for public and private sets; (2) for each perturbation type; (3) for each model; (4) for the previously published and unpublished problems.
For the final report, we will use nonparametric bootstrap confidence intervals over test examples and paired significance testing (e.g., McNemar’s test on per-example decisions) for the comparisons of the final submissions.

Both in-competition and final evaluations will be run on the isolated infrastructure of organizers, with 32 CPU cores, 512GB of RAM and eight large-scale N GPUs for both in-competition and the final evaluation. We will use the submission bundle interface predefined by the Codabench competition platform\footnote{\url{https://codabench.org}}, assuring reproducibility of valid submissions in docker containers.


%
\subsection{Baselines, code, and material provided}

\paragraph{Baselines}

For the competition, we will implement at least three baselines provided as starting bundles: (1)~a simple probing classifier \citep{10.1162/coli_a_00422,waldis-etal-2024-holmes} predicting robustness label from the internal representations of last input token on the best-performing layer, (2)~a classifier based on model-internal confidence signals during chain-of-thought generation \citep{grunefeld2026tracinguncertaintylanguagemodel}, and (3)~a data attribution method \citep{park2024data} assessing robustness as the impact and scale of memorization.
We have already implemented the first two of the baselines using different types of classifiers and evaluated them with DeepSeek-R1-0528-Qwen3-8B\footnote{DeepSeek-R1-0528-Qwen3-8B was among the three highest-ranked models in AIMO 3.} on our sample validation set in stratified 10-fold cross-validation. We found that (1) the probing classifier and uncertainty-based classifier achieves up to \emph{58.37} (±7.5) and \emph{69.23\%} (±10.13) of accuracy, respectively. As such, both baselines outperform the random-guessing baseline (50\%), showing that our newly proposed task is both \emph{feasible} as models' internals exhibit an informative signal, and \emph{challenging} as the results leave plenty of room for improvements.

All these baselines will be provided as implementations on the competition GitHub\footnote{\url{https://github.com/aimo-interp/baselines}} with instructions how to build them into a ready-to-submit Codabench submission bundle.
As such, participants will start their implementations from functional prototypes, minimising their time spent in familiarizing with the formatting of submissions.

Each baseline will publish its full lifecycle codebase, including reusable competition data loaders, training code, local evaluation, submission bundle build, and submission interface.
We will provide all implementations, together with a high-level walkthrough of these steps, as a starting guide in this GitHub's README by the announcement of the competition (July 1, 2026).
The Starter kit is already referenced from the header of the competition website.

Beyond this Starter kit, we already make available:

\begin{enumerate}[leftmargin=1.5em]
    \item Public, human-readable sample of train and validation sets\footnote{\url{https://huggingface.co/datasets/aimo-interp/val-sample}} with examples in a format defined in Section~\ref{sec:tasks}, i.e. containing a collection of samples with (1) a math word problem, (2) a model~(i.e. a HuggingFace model ID) and (3) a label of the tested distribution shift;
    \item A ready-to-run implementation of a baseline competition systems\footnote{\url{https://github.com/aimo-interp/baselines}};
    \item Evaluation codebase that will exactly match the one ran on a private test set on our servers through Codabench\footnote{\url{https://github.com/aimo-interp/evaluation}}; All model types included in the private test set will be covered in the public validation set;
    \item A convenient Codabench competition interface, including a leaderboard presenting the results of evaluations of participants' submissions on the private test set in real time.\footnote{Codabench interface of the competition is now privately available at \href{https://www.Codabench.org/competitions/16180/?secret_key=cb3b2ef1-7be5-4a93-a5ca-d95e5c535863}{\underline{this url}}.}
\end{enumerate}

All these resources will be linked from the competition's Starter kit. Possible rough edges will be refined via testing submissions before the competition opens and in the Warmup competition phase.


%
%
\subsection{Website, tutorial and documentation}
The public challenge website, \url{https://aimo-interp.github.io}, already presents the challenge motivation, task, tracks, environment, a concise ``How to participate'' guide, FAQ covering tracks, data usage, hardware access, and submission format, a dedicated organiser contact address, and a timeline overview with key competition dates.


%
\section{Organizational aspects}

\subsection{Protocol}
The competition will run fully online before the conference. Each team will complete these steps:
\begin{enumerate}[leftmargin=1.5em]
    \item Register on Codabench, read the rules, and download the submission bundle with a baseline. Full public split of validation data will be publicly available and referenced from the website;
    \item (optional) Apply for the compute resources provided by the organisers via the Fields Model Initiative\footnote{\url{https://www.fieldsmodel.org}} by completing the online form;
    \item Submit executable containers on Codabench. These will be evaluated on the private test set on our servers with no Internet connection and fixed compute budget (1 hour per problem on 8-GPU nodes);
    \item After the competition deadline, participants will be required to submit a 1--2-page technical report, summarizing the methods applied in their submission.
\end{enumerate}

Cheating and overfitting will be limited through hidden tests, limits on the number of submissions, offline execution, and compliance checks after the deadline.
Before the competition start, the submission interface and leaderboards will be tested with the organisers' baselines.

We support the openness of the competition to underresourced communities not only by an open call for application for compute resources (§\ref{sec:res_provided}), but also by introducing a dedicated small-model track (§\ref{sec:tasks}) and a standardised evaluation pipeline running on our hardware.

\subsection{Rules and Engagement}

\paragraph{Submission rules.}
Submissions must follow the following rules\footnote{As of 15 May 2026. Please check \url{https://aimo-interp.github.io} for updates.} to prevent overfitting:
\begin{enumerate}[leftmargin=*]
    \item \textbf{Data:} teams may not use extra labeled data for the same perturbation types as in our final validation set; however, unlabeled data and labeled data for other perturbations are allowed.
    \item \textbf{Signals:} teams can use weights, activations, token probabilities, or any other black-box statistics as inputs for their methods.
    \item \textbf{Tracks:} teams may submit to the Main or Small models track, but each submission will count only for the chosel submission track.
    \item \textbf{Report:} ranked teams must submit a technical report within one week for compliance checks.
\end{enumerate}
These rules discourage overfitting to our robustness-labeling mechanism, reward methods' generalization, and keep participation open across affiliations, geographies, career and technical levels.



\noindent \textbf{Communication}
All official competition updates will be communicated via email to all participants enrolled in the competition in Codabench and in the competition's Discord server on the announcements channel.
Participants can always reach out to organizers by using the official email (aimo-interp@gmail.com) or by tagging one of the organizers in the Discord channel.
The official competition website will maintain and update a dedicated FAQ section with participants' questions.


\subsection{Schedule and readiness}
\noindent \textbf{Schedule\ \ }\textbf{1) By July 15:} finalize splits, baselines, and Codabench setup;
\textbf{2) Late July:} competition announcement, release of validation data and Starter bundles with baselines;
\textbf{3) By Aug. 2026:} submissions warm-up phase permitting minor changes in the interface;
\textbf{4) Aug. 1--Nov. 1:} main competition phase;
\textbf{5) Oct. 1:} final submission deadline;
\textbf{6) Nov. 15:} technical reports due;
\textbf{7) Nov. 15:} results validation and analysis;
\textbf{8) Dec. 1:} final leaderboard and organizers report publications, workshop materials and schedule finalized;
\textbf{9) Dec. 11/12:} competition workshop.
The schedule gives participants at least \emph{three months} from the announcement to final evaluation, which is suitable for training or designing potentially complex interpretability methods.

\noindent \textbf{What is already ready\ \ }
We have a near-final version of the public website, including task specifications, contacts, and FAQs. Further, we already have implementations of evaluations, a competition dataset sample, Codabench competition and implementations of baselines. We also have agreements on data and resource sharing within AIMO and the Fields Model Initiative allowing us to immediately provide the compute resources to interested participants.
An outstanding task that we are working on now is a further extension of our datasets with the non-public problems from AIMO and JMO.

\noindent \textbf{Contingency plan}
In the case that fewer than 3 participating teams will submit by the deadline, we will implement further baselines from the prior work and include them in the reports, delivering on our scientific goals.
In the case that the allocated compute resources for our evaluations and participants become unexpectedly unavailable, we may restrain the Small models track towards even smaller models and provide the resources for evaluation, including GPUs, from commercial sources.
Our competition will involve a private sets of problems from AIMO that are not publicly available. In the unlikely case that AIMO organisers withdraw from this agreement, we will replace the test set with unpublished problems of comparable complexity from JMO.
We will move forward with the competition according to the schedule, regardless of these externalities.

%
\subsection{Competition promotion and incentives}

We will promote the competition through the organizers' research networks, including both AI and less-represented pure mathematics networks. We will also advertise the call for participation through official competition social-media accounts and reposts by organisers on Bluesky, LinkedIn, and X.
We will also use our networks to promote the challenge through online channels and in-person presentations at closely related venues taking place during the competition, including the GEM workshop at ACL 2026, the Mechanistic Interpretability and AI4Math workshops at ICML 2026, and the BlackboxNLP workshop at EMNLP 2026. These venues will help us reach a broad range of researchers interested in evaluation, interpretability, reasoning, and mathematical problem solving.

\paragraph{Incentives}
All participants will be invited to submit a non-archival technical report after the competition deadline. Submissions with strong empirical results or broadly impactful methodological lessons documented in their report will be invited to present at the in-person competition venue. As of today, we are awaiting confirmation of industrial sponsorship for a 500 USD award for the best-ranked method in each track.

\subsection{Competition track workshop and dissemination}

We will create and maintain a dedicated Discord server where we will encourage team formation, discussion of ideas and preliminary results, and allow participants to reach out to us directly with any questions.
The competition workshop will allow submitting a non-archival one-page technical report for each submission.

Based on the reports, we will invite selected participants to present their methods on site at the workshop, highlighting strong results and generalizable methodological lessons.
The workshop will include the overview presentation with the final results, general takeaways, summarizing what worked and what did not.
The program will also include 2--3 invited talks from invitees or members of the advisory board, featuring experts in evaluation, interpretability and generalization.

With the participants' permission, we will also promote the findings documented in the reports on the competition's social media accounts.
We will release the pre-print with the competition findings before the workshop's in-person event.

\section{Resources}

\subsection{Organizing team}
The organizing team combines expertise in evaluation, interpretability and data collection with top-tier mathematicians, setting us to a unique position for curating a high-quality data collection that our competition builds upon. Close involvement of organizers of AIMO, ranking among the 30 largest competition in the history of Kaggle, also brings to our team a unique expertise with competition organization at scale.

In overview, our team consists of the following experts and their roles:

\begin{itemize}[leftmargin=1.5em]
    \item \textbf{Michal \v{S}tef\'anik} (National Institute of Informatics, Japan) -- Competition coordination, website maintenance and communication
    \item \textbf{Philipp Mondorf} (MCML / MaiNLP LMU) -- Competition coordination, baseline implementations, leaderboard
    \item \textbf{Andreas Waldis} (University of T\"ubingen) -- Baseline implementations and beta testing
    \item \textbf{Qianying Liu} (National Institute of Informatics, Japan) -- Olympiad math expert: validation of robustness evaluations
    \item \textbf{Chuan Yang} (Fuzhou University) -- Olympiad math expert (top-100 in Chinese National Olympiads): supervision and coordination of data collection with mathematical communities
    \item \textbf{Michal Spiegel} (Masaryk University) -- Data: robustness labels collection
    \item \textbf{Robert Belanec} (Faculty of Information Technology, Brno University of Technology / Kempelen Institute of Intelligent Technologies) -- Data curation, robustness labels, compute allocation. 
    \item \textbf{Josef Kuchař} (Masaryk University) -- Codabench \& infrastructure development
    \item \textbf{Marek Kadlčík} (Masaryk University) -- Data: Symbolic chains and permutations design
    \item \textbf{Adam Vawda-Oomerjee} (National Institute of Informatics / University College London) -- Infrastructure support, Codabench administration, beta testing
    \item \textbf{Chaoran Liu} (National Institute of Informatics, Japan) -- Infrastructure: participants' support
    \item \textbf{Simon Frieder} (AIMO Manager / University of Oxford / Benchmarks \& Baselines) -- Advisory board: Competition infrastructure design, coordination on accessing AIMO models and data
    \item \textbf{Barbara Plank} (MCML / MaiNLP LMU) -- Advisory board: evaluation and data collection
    \item \textbf{Fazl Barez} (University of Oxford) -- Advisory board: Interpretability methods
    \item \textbf{Pontus Stenetorp} (University College London / National Institute of Informatics, Japan) -- Advisory board: evaluation and adversarial data collection, initiative promotion
\end{itemize}


\subsection{Resources provided by organizers}
\label{sec:res_provided}

The competition will provide resources and infrastructure, opening the abilities of frontier industrial labs to the broader community:
\begin{itemize}[leftmargin=1.5em]
    \item \textbf{Large-scale compute:} participating teams may request an access to up to 10,000 H200 GPU-hours of compute, justified in the submitted proposal;
    \item \textbf{Frontier reasoning models:} we will provide unrestricted access to best-performing AI systems from AIMO in standalone, fully reproducible containers;
    \item \textbf{Standardized evaluation infrastructure:} official submissions will be evaluated on our infrastructure with access to eight large-scale GPUs, supporting even cost-heavy inference methods;
\end{itemize}


\subsection{Support requested}
We would appreciate standard support from the NeurIPS 2026 Competition Track organizers, especially inclusion in official promotion channels to help us reach relevant communities, as well as appropriately sized on-site meeting space for the competition session.
We would also welcome any additional guidance from the Competition Track organizers to support the smooth and successful execution of the competition.

\begin{ack}

This work was partially funded by the European Union, under the project LorAI - Low Resource Artificial Intelligence, GA No. \href{https://doi.org/10.3030/101136646}{101136646}.

Part of the validation and test set data were obtained using computational resources funded by the European Union – NextGenerationEU through the Recovery and Resilience Plan for Slovakia under Project No. 17I03-04-P02-00001: Development and Construction of a Supercomputer for the National Supercomputing Center.

\end{ack}

\bibliographystyle{plainnat}
\bibliography{bibliography}

\appendix

\section{Biography of all team members}
\label{sec:biography}

\paragraph{Michal \v{S}tef\'anik} (Lead organizer) is a Specially Appointed Researcher at the National Institute of Informatics, Japan (NII) with a track record in evaluation, interpretability and generalization of large language models. Prior to NII, he was a Postdoctoral Research Associate at the University of Helsinki, Finland and a PhD Researcher at Masaryk University, Czechia. He has a track record of first-authored publications at *ACL and EMNLP conferences focused on the robustness of neural language models. His research in this area was awarded several prizes and awards, including two Best Paper awards and a Dean's Award (top 10\%) for his dissertation studying covariates of robustness of neural language models. Beyond research, Michal holds over 8 years of AI research experience across startups and corporates.

\paragraph{Philipp Mondorf} (Core organizer) is a third-year PhD student at LMU Munich specializing in reasoning, evaluation, interpretability, and generalization in language models. He has an established publication record at leading AI and NLP venues, including ICLR, ICML, ACL, EMNLP, EACL, and COLM. During his time at Meta FAIR, he worked on improving the reasoning robustness of LLMs in collaboration with leading experts in the field, including Dieuwke Hupkes and Jesse Dodge. He is currently a visiting researcher at NYU and Princeton University, where he conducts research on compositional generalization with Brenden Lake and Todd Gureckis.

\paragraph{Andreas Waldis} (Core organizer) is a PostDoc and Junior Group Lead at the University of Tübingen. His research focuses on understanding and improving the reliability of language models, with an emphasis on methods that integrate internal and behavioral perspectives. His work spans evaluation, interpretability, and generalization, with applications to computational argumentation and societal alignment.
He co-organized the first shared task on Perspective Argument Retrieval (ArgMining@ACL2024), the previous iteration of INTERPLAY in 2025, and KnitTogether25, a European summit supporting early-career researchers in computational linguistics, and regularly serves as an Area Chair at ARR.

\paragraph{Qianying Liu} is a Specially Appointed Researcher at the Large Language Model Research and Development Center of National Institute of Informatics, Japan (NII). Her research focuses on multilingualism, reasoning, and interpretability of large language models, with publications in major venues including ACL, EMNLP, TASLP, and AAAI. Prior to joining NII, She completed PhD studies at Kyoto University under the supervision of Professor Sadao Kurohashi. She has also regularly served as a Senior Area Chair for ARR and various other major conferences.

\paragraph{Chuan Yang} is an Associate Professor in the School of Mathematics and Statistics at Fuzhou University. Her research focuses on discrete modeling, combinatorial, and continuous optimization. Her research works have been accepted or published in SIMAX, Sci. China Math., and Commun. Math. Sci. Prior to joining Fuzhou University, she received her PhD in Computational Mathematics from Peking University.

\paragraph{Adam Vawda-Oomerjee} is a Visiting Researcher at the Large Languge Model Research and Development Center at the National Institute of Informatics, Japan (NII), and a Research Assistant in the Natural Language Processing Group (NLP) at University College London (UCL). His work focuses on understanding reasoning, reinforcement learning, and world models. 

\paragraph{Michal Spiegel} is a student researcher at Masaryk University, Czechia, and a Machine Learning Engineer at Filevine. Previously, he served as an NLP researcher at the Kempelen Institute of Intelligent Technologies. Specializing in mechanistic interpretability, his research focuses on algorithmic reasoning and length generalization within large language models. He has an established early-career track record with publications at major NLP venues, including ACL and EMNLP. Recently, he served as a visiting researcher at the National Institute of Informatics (NII), Japan, where he focused on robustness evaluations in mathematical reasoning.

\paragraph{Robert Belanec} is a PhD student at the Brno University of Technology and Kempelen Institute of Intelligent Technologies, focusing on adaptation of LLMs using parameter-efficient multi-task transfer learning with limited labeled data. Moreover, his research interests also include generalizability, interpretability of AI models, and model-merging methods.

\paragraph{Josef Kuchař} is a student researcher at Masaryk University, Czechia. He is interested in artificial intelligence agents and software engineering, with a particular focus on alternative LLM architectures. He co-authored or first-authored four publications presented at ACL, EMNLP, and LREC conferences. His work combines practical software engineering with machine learning research. In this competition, he will contribute his engineering expertise to design and develop a suitable Codabench interface.

\paragraph{Marek Kadl\v{c}\'ik} is a second-year PhD student at Masaryk University, Czechia, specialized in reasoning, reinforcement learning, and the generalization of neural models. He has an established publication record at leading NLP venues such as ACL and EMNLP, and his prior work in reinforcement learning have been honored with two Dean’s Awards and one Rector's Award. Beyond academia, Marek has several years of practical experience in R\&D spanning applications in NLP, computer vision, and reinforcement learning.

\paragraph{Chaoran Liu} is currently a Project Associate Professor at the National Institute of Informatics (NII), Japan. Previously, he was a Research Scientist at RIKEN and a Specially Appointed Assistant Professor at Osaka University. His recent work centers on large language models, with a focus on pretraining analysis, data attribution, training dynamics, and formal mathematical reasoning using LLM-based agents. His broader research background includes signal processing, robotics, and human–robot interaction. In this competition, he contributes infrastructure expertise and participant support.

\paragraph{Simon Frieder} is the AIMO Prize Manager and founder of Benchmarks \& Baselines the non-profit, dedicated to benchmarks LLMs on reasoning tasks. Previously, he studied toward a PhD at the University of Oxford, UK. He holds separate degrees in mathematics and computer science, and his research has been featured in popular science and technology outlets such as Ars Technica and ZDNet, as well as cited in AI-related reports by the German government. He has published at leading machine learning conferences, including NeurIPS, ICML, and ICLR. He will advise on organizational aspects and assist with coordination in sharing findings and resources with the AIMO competition.

\paragraph{Barbara Plank} is a Full Professor and Chair for AI and Computational Linguistics at LMU Munich, where she leads the Munich AI \& NLP Lab (MaiNLP) and co-directs the Center for Information and Language Processing (CIS). Her research focuses on natural language processing, especially robust and inclusive models that account for domain shift, language variation, limited supervision, and human label variation, with broader interests in interpretability, reasoning, and trustworthy evaluation. In this competition, she will advise on robust NLP evaluation and methods for handling data variation and annotation uncertainty.

\paragraph{Fazl Barez} is a Senior Researcher at the University of Oxford, where he is Principal Investigator of the Technical Safety \& Governance (TSG) Lab and Technical Director of the AI Governance Initiative. His group works across AI safety, interpretability, and technical governance. At Oxford, he teaches the AI Safety and Alignment course, and alongside his academic work, he is Principal Scientist at Martian, where they work on understanding machine intelligence. His research is supported by OpenAI, Anthropic, Schmidt Sciences, NVIDIA, and others. In this competition, he will advise on support for interpretability methods.

\paragraph{Pontus Stenetorp} is a Professor of Natural Language Processing (NLP) at University College London (UCL), Deputy Director of the UCL Centre for Artificial Intelligence and Specially Appointed Professor at National Institute of Informatics, Japan. He leads the UCL NLP group and his primary research contributions and interests lie in the areas of model and system development, evaluation, and analysis. His research has received awards at leading conferences in his field: outstanding paper at EACL 2017, best paper at EACL 2021, outstanding paper at ACL 2022, and best paper ACL 2023. To date he has published more than 100 publications, which have been cited more than 9,000 times. In this competition, he will advise on adversarial data collection methodologies.

\section{Perturbation types and adversarial evaluations} 
\label{appx:adversarial_evaluations}

Following GSM-Symbolic~\citep{mirzadeh2025gsmsymbolic} and related symbolic-template approaches~\citep{mondorf2026lpdsevaluatingllmrobustness}, we evaluate robustness under answer-preserving and adversarial perturbations of annotated reasoning-chain examples. The open list of perturbations that we implemented to date of submission is:
\begin{itemize}
    \item \textbf{Rephrase}: semantic-preserving rewrites of the original problem.
    \item \textbf{Rename}: renaming of variables or entities while preserving the underlying structure.
    \item \textbf{Domain}: transfer of the same mathematical structure to a different surface domain.
    \item \textbf{Distract}: insertion of irrelevant but plausible information.
    \item \textbf{Typos}: introduction of harmless typographical or formatting noise.
    \item \textbf{Expert perturbations}: answer-preserving adversarial edits making use of the expert knowledge of the problem (mathematical conventions, unstated premises or assumptions).
    \item \textbf{Expert no-solution}: unanswerable variants of the problems (e.g. missing some of the problem premises) intended to test unsolvability detection.
\end{itemize}

\begin{table}[h]
\centering
\footnotesize
\setlength{\tabcolsep}{6pt}
\begin{tabular}{llll}
\toprule
AIMO ref. problem & Model & Perturbation type & Accuracy drop \\
\midrule
\texttt{bbd91e} & Qwen3.5 & \texttt{expert no-solution} & $100\% \to 10\%$ ($90\%$) \\
\texttt{a1d40b} & Qwen3-8B & \texttt{expert no-solution} & $100\% \to 10\%$ ($90\%$) \\
\texttt{71beb6} & GPT-OSS-120B & \texttt{expert perturbations} & $70\% \to 20\%$ ($50\%$) \\
\texttt{057f8a} & Qwen3.5 & \texttt{distract} & $60\% \to 20\%$ ($40\%$) \\
\texttt{057f8a} & Qwen3.5 & \texttt{domain} & $60\% \to 10\%$ ($50\%$) \\
\texttt{057f8a} & Qwen3.5 & \texttt{rename} & $60\% \to 20\%$ ($40\%$) \\
\texttt{480182} & GPT-5.2 & \texttt{rephrase} & $100\% \to 60\%$ ($40\%$) \\
\texttt{057f8a} & Qwen3.5 & \texttt{typos} & $60\% \to 30\%$ ($30\%$) \\
\texttt{349493} & Gemini-3.1 Pro & \texttt{typos} & $60\% \to 30\%$ ($30\%$) \\
\midrule
\end{tabular}
\caption{Examples of paired accuracy drops between original problems and perturbed variants generated from symbolic chains. Problems from the AIMO public reference set.}
\label{tab:drops}
\end{table}

Table \ref{tab:drops} shows performance drops of frontier AI models on permutations of AIMO 2 problems obtained using the annotations of symbolic reasoning chains collected for the purpose of this competition.
These evaluations show that our symbolic chains allow for generating perturbations able to disrupt models' predictions and identify their non-robust reasoning scenarios.

\end{document}